\documentclass[sigconf,nonacm,natbib=false,balance=false]{acmart}
\usepackage{amsmath}
\usepackage{booktabs}
\usepackage{url}
\setcopyright{none}
\hypersetup{pdftitle={Pre-training with Graph Transformers},
  pdfauthor={Jiaming Wang, Thomas Laurent, Xavier Bresson},
  pdfkeywords={graph neural networks, graph transformers, pre-training}}

\begin{document}
\title{Pre-training with Graph Transformers}
\author{Jiaming Wang}
\email{jamie-w@nus.edu.sg}
\affiliation{\institution{National University of Singapore}\country{Singapore}}
\author{Thomas Laurent}
\email{tlaurent@lmu.edu}
\affiliation{\institution{Loyola Marymount University}\city{Los Angeles}\country{USA}}
\author{Xavier Bresson}
\email{xaviercs@nus.edu.sg}
\affiliation{\institution{National University of Singapore}\country{Singapore}}

\renewcommand{\shortauthors}{Wang et al.}
\begin{abstract}
This article investigates pre-training strategies for graph transformers in the biochemistry domain. By conducting comprehensive experiments, the study reveals that supervised pre-training using computed properties as labels provides the highest performance gain on downstream tasks. The results also highlight the importance of constraining model capacity to mitigate overfitting in graph transformers.\footnote{Project code: \url{https://anonymous.4open.science/r/pretrain-graphtransformer}}
\end{abstract}
\maketitle

\section{Introduction}
Pre-training large transformer-based models \cite{ref15} using extensive datasets has yielded significant advancements in diverse fields including computer vision \cite{ref4}, \cite{ref8} and natural language processing \cite{ref3}, \cite{ref2}. Nevertheless, the feasibility of applying pre-training techniques to transformer-based graph neural networks (GNNs) using graph data remains an unresolved issue due to several challenges associated with transfer learning in this context. These challenges encompass sparsity of graph connections, limited availability of large-scale datasets, and the diversity observed across different graph domains. Furthermore, unlike text and image data, which often possess discernible `locality' characteristics, graph data typically lack such inherent notions of locality. Consequently, when pre-training transformer-based GNNs, special consideration must be given to account for those constraints.

Several studies have proposed different transformer-based graph neural networks \cite{ref11}, \cite{ref21}, \cite{ref6}, \cite{ref22}, \cite{ref23}, \cite{ref10}, \cite{ref19}, \cite{ref13} that have significantly enhanced the performance of the original graph transformers \cite{ref5} across various tasks. However, these works have primarily focused on the performance of GNNs in specific tasks and have not emphasized the pre-training setting, which requires the acquisition of generalized representations for downstream tasks.

Hu et al.\ \cite{ref9} investigated pre-training strategies for GIN \cite{ref20}, and indicated negative transfer when pre-training with attention-based GAT \cite{ref16}. Conversely, our own experimental findings suggest that attention-based models are susceptible to overfitting, and we can mitigate this issue by constraining the model capacity or extending the size of the pre-training dataset. While other studies such as \cite{ref12} and \cite{ref24} have demonstrated that pre-training transformer-based GNNs can enhance their performance on downstream tasks, their primary focus lies in a specific self-supervised training strategy rather than determining the optimal strategies for pre-training graph transformers. In contrast, our work aims to address the question of identifying the most effective pre-training strategies for graph transformers \cite{ref5}.

To address this inquiry, we conducted comprehensive experiments by first pre-training a graph transformer model using different strategies and then evaluating the pre-trained model on various downstream tasks in the biochemistry domain. Remarkably, our experimental outcomes indicate that the most favorable performance improvement, reaching an average increase of 5.8\% in ROC-AUC, is achieved through supervised pre-training employing computed properties as labels, as opposed to biochemical assays. This finding stands in contrast to previous research \cite{ref9}, which suggested the utilization of self-supervised learning alone or in combination with supervised learning utilizing laboratory assays as targets. Furthermore, we discovered that, as a more expressive architecture, graph transformers require more constraints on model capacity to counter overfitting, particularly in situations with limited data availability.

\section{Experimental Setup}
Our study aims to identify the key factors that significantly impact the performance of downstream tasks in the context of pre-training graph transformers. To achieve this objective, we conducted a series of comprehensive experiments employing the following setups.

\subsection{Datasets}\label{sec:datasets}
We utilized three biochemistry datasets of varying sizes and labels for pre-training: ZINC 250K, ZINC 1M \cite{ref14}, and ChemBL \cite{ref7}. These datasets consist of 250,000, 1 million, and 430,000 molecules, respectively. Each graph in the datasets represents a single molecule. We use a single atom type (atomic number) as the node feature. The edge feature consists of a single bond type (single bond, double bond, triple bond or aromatic bond). The ZINC dataset does not possess any inherent labels, whereas ChemBL encompasses 1,310 diverse and extensive biochemical assays.

Additionally, we computed a collection of molecular properties for each molecule using RDkit\footnote{\url{http://www.rdkit.org/}}, which served as the regression targets for supervised learning. Two experimental settings were explored: utilizing all 127 available properties\footnote{\url{https://www.rdkit.org/docs/GettingStartedInPython.html\#list-of-available-descriptors}} (denoted as \textbf{SLCP-ALL} in the experiment result), as well as a smaller selection of properties (denoted as \textbf{SLCP-DD}) commonly employed in drug discovery, namely SAS, QED, and LogP. By incorporating these computed properties, our aim was to enable the network to identify pertinent molecular features, such as the count of aromatic rings, in order to enhance performance in downstream tasks. It is worth noting that this approach differs from previous works \cite{ref12}, \cite{ref24}, \cite{ref9} that primarily focused on self-supervised pre-training or utilized ``natural'' targets, which involve laborious and resource-intensive biological experiments in wet laboratories.

To evaluate the pre-trained model, we employed eight datasets from MoleculeNet \cite{ref18} as downstream tasks. It is a curated benchmark for molecular property prediction (binary classification) within the biochemistry domain. Following \cite{ref9}, We split the downstream data using scaffold split to evaluate the model's ability to generalize to out-of-distribution scenarios.

\subsection{GNN architectures}
The original implementation of Graph Transformer \cite{ref5} was employed in this study, as it represents a direct extension of the transformer model to graphs. Three different model sizes were tested, characterized by varying numbers of hidden units: 304 (6.6 million parameters, denoted as \textbf{GTL}), 168 (2 million parameters, denoted as \textbf{GTM}), and 96 (0.7 million parameters, denoted as \textbf{GTS}). All models consisted of 5 GNN layers and 8 attention heads. The READOUT function utilized average pooling.

Both node features (atom type) and edge features (bond type) were used as inputs in our experiments. In each experiment, the model underwent pre-training on datasets described in section~\ref{sec:datasets} for a fixed duration of 100 epochs, followed by fine-tuning on downstream tasks for 40 epochs.

Our study compared three different positional encoding schemes: Laplacian positional encoding (\textbf{LapPE}) \cite{ref1}, \cite{ref5}, random walk positional encoding (\textbf{RWPE}) \cite{ref5}, and the absence of any positional encoding (\textbf{NoPE})

Furthermore, we conducted additional tests using GIN with the implementation described in \cite{ref9}. Notably, the large graph transformer model (GTL) shared the same number of hidden units as GIN, while the medium-sized graph transformer model (GTM) had approximately equivalent parameter count. In contrast, the smallest graph transformer model (GTS) had roughly one third of the parameter count compared to GIN.

\subsection{Pre-training Strategies}\label{sec:strategies}
We considered five different pre-training strategies:
\begin{itemize}
\item Supervised learning with biochemical assays (denoted as \textbf{SLBA}), in which we use the biochemical assays as the target label during pre-training. Note this is the most common setup in previous studies;
\item Supervised learning with computed properties (denoted as \textbf{SLCP-ALL} and \textbf{SLCP-DD}), as described in section~\ref{sec:datasets}. This strategy has not been extensively studied in previous works;
\item Self-supervised masking node attributes (denoted as \textbf{MASK}), in which we randomly masked 30\% of nodes similarly to \cite{ref3};
\item Self-supervised DGI \cite{ref17} (denoted as \textbf{DGI}),
\item Combining supervised learning with self-supervised learning by 1) linearly combining the supervised and self-supervised loss, which is denoted as \textbf{DGI+$\lambda$SL},
\[
\mathcal{L}_{\text{combined loss}} = \lambda\,\mathcal{L}_{\text{supervised loss}} + \mathcal{L}_{\text{DGI loss}}
\]
2) self-supervised training followed by supervised training, as proposed in \cite{ref9}, denoted as \textbf{DGI-SL}
\end{itemize}

\section{Results}
The results of our experiment are summarized in Table~\ref{tab:results}. Our analysis of the experimental results reveals the following insights:

\textbf{(1)} Among various pre-training strategies described in Section~\ref{sec:strategies}, pre-training with supervised learning using computed properties shows the highest performance gain on downstream tasks. This pattern is observed in both the GT and GIN models. We hypothesize that the dense computed properties enable the neural network to learn useful and general features of molecules, which prove beneficial for downstream tasks. This finding has significant implications in the computational biochemistry domain, as we can easily obtain a large number of ``artificial'' targets using various tools on existing extensive datasets containing millions of molecules \cite{ref14}, \cite{ref7}. Our experiment results suggest that performing supervised pre-training using such datasets can greatly enhance performance on tasks with limited data. However, the indiscriminate use of all available computed properties may hinder performance, as indicated by experiment \#28. Consequently, the selection of the most suitable targets for pre-training remains an open question.

\textbf{(2)} By comparing the average test ROC-AUC of graph transformers with the same hyperparameters but pre-trained using different datasets, or with different numbers of parameters but pre-trained using the same datasets, we conclude that graph transformer models are prone to overfitting and therefore benefit from constraining the model capacity when the available data is limited. Notably, the GT model with one-third the number of parameters of GIN, pre-trained on the ZINC 1M dataset, exhibits the highest performance gain on downstream tasks (experiment \#19). We hypothesize that this phenomenon occurs because GTs are more flexible models, thus requiring stronger regularization compared to GIN. This finding aligns with previous empirical studies on transformer-based models.

\textbf{(3)} Self-supervised pre-training using DGI and Masking learning objectives yield similar results. Moreover, combining self-supervised learning with supervised learning, as suggested in \cite{ref9}, does not yield the best performance gain. We hypothesize that this is because using the dense computed properties (SLCP) provides a better training signal than the sparse biochemical assays (SLBA) used in previous works.

\textbf{(4)} Random walk positional encoding (RWPE) demonstrates superior performance compared to LapPE and NoPE. This may be attributed to the requirement of randomly flipping signs in LapPE, which necessitates a greater number of training epochs.

\begin{table*}[t]
\centering
\small
\setlength{\tabcolsep}{4pt}
\renewcommand{\arraystretch}{1.0}
\begin{tabular}{rlllrrrrrrrrr}
\toprule
& \multicolumn{3}{c}{PRE-TRAINING METHODS} & \multicolumn{9}{c}{DOWNSTREAM TASKS} \\
\cmidrule(lr){2-4}\cmidrule(lr){5-13}
ID & GNN & Method & Dataset & CLINTOX & MUV & HIV & SIDER & BACE & TOX21 & BBBP & TOXCAST & \textbf{AVERAGE} \\
\midrule
1 & GTL - LapPE & - & - & 54.2 & 51.7 & 53.3 & 51.0 & 59.8 & 58.4 & 69.9 & 55.5 & 56.7 \\
2 & GTM - LapPE & - & - & 55.8 & 52.2 & 56.3 & 56.2 & 69.8 & 61.2 & 72.6 & 57.1 & 60.2 \\
3 & GTL - RWPE & - & - & 66.3 & 56.1 & 61.9 & 55.0 & 50.9 & 58.1 & 72.5 & 56.5 & 59.7 \\
4 & GTM - RWPE & - & - & 65.1 & 51.8 & 59.6 & 54.1 & 67.0 & 63.6 & 72.3 & 61.1 & 61.8 \\
5 & GTL - NoPE & - & - & 64.7 & 52.1 & 61.2 & 51.1 & 63.8 & 52.8 & 68.8 & 59.3 & 59.2 \\
6 & GTM - NoPE & - & - & 61.2 & 52.4 & 60.0 & 54.3 & 62.8 & 59.7 & 75.2 & 56.6 & 60.3 \\
7 & GTS - NoPE & - & - & 63.4 & 54.2 & 61.4 & 58.5 & 67.3 & 64.8 & 77.8 & 60.0 & 63.4 \\
8 & GIN & - & - & 60.3 & 73.1 & 73.0 & 56.0 & 71.2 & 74.4 & 76.0 & 63.2 & 68.4 \\
\midrule
9 & GTM - RWPE & MASK & ZINC 250K & 60.4 & 59.3 & 61.2 & 54.3 & 64.0 & 66.9 & 76.0 & 61.0 & 62.9 \\
10 & GTM - RWPE & DGI & ZINC 250K & 59.4 & 54.4 & 62.3 & 54.0 & 66.5 & 64.7 & 74.7 & 60.7 & 62.1 \\
11 & GTL - LapPE & MASK & ZINC 250K & 65.9 & 52.9 & 54.0 & 53.0 & 60.4 & 52.6 & 70.2 & 57.7 & 58.3 \\
12 & GTL - LapPE & DGI & ZINC 250K & 62.2 & 48.9 & 56.4 & 54.8 & 56.6 & 55.5 & 71.2 & 54.9 & 57.5 \\
13 & GIN & MASK & ZINC 1M & 60.3 & 70.1 & 73.0 & 56.0 & 71.2 & 74.4 & 76.0 & 63.2 & 68.0 \\
\midrule
14 & GTM - RWPE & SLCP-DD & ZINC 250K & 72.0 & 60.0 & 70.1 & 56.9 & 67.6 & 67.1 & 79.9 & 62.1 & 67.0 \\
15 & GTL - RWPE & SLCP-DD & ZINC 250K & 59.9 & 56.9 & 64.6 & 54.9 & 55.7 & 59.8 & 68.0 & 56.5 & 59.5 \\
16 & GTL - NonePE & SLCP-DD & ZINC 250K & 63.0 & 52.3 & 59.4 & 60.2 & 62.5 & 62.3 & 73.5 & 53.0 & 60.8 \\
17 & GTM - NoPE & SLCP-DD & ZINC 250K & 65.1 & 58.7 & 67.4 & 58.5 & 65.9 & 68.5 & 75.4 & 60.0 & 64.9 \\
18 & GTS - NoPE & SLCP-DD & ZINC 250K & 61.3 & 68.2 & 77.4 & 57.6 & 70.3 & 73.5 & 78.1 & 61.0 & 68.4 \\
19 & GTS - RWPE & SLCP-DD & ZINC 1M & 69.4 & 73.4 & 75.5 & 57.5 & 76.3 & 74.8 & 78.8 & 61.6 & \textbf{70.9} \\
20 & GTM - LapPE & SLCP-DD & ZINC 250K & 64.9 & 50.4 & 64.8 & 57.7 & 70.0 & 69.1 & 79.0 & 62.4 & 64.8 \\
21 & GTL - LapPE & SLCP-DD & ZINC 250K & 62.1 & 53.5 & 59.0 & 53.5 & 57.0 & 55.1 & 68.9 & 54.5 & 57.9 \\
22 & GTL - LapPE & SLCP-DD & ZINC 1M & 66.3 & 62.6 & 65.0 & 58.3 & 71.4 & 64.9 & 76.5 & 59.2 & 65.5 \\
23 & GTM - RWPE & SLCP-DD & ZINC 1M & 67.9 & 65.0 & 76.2 & 55.0 & 75.1 & 72.2 & 81.1 & 64.3 & 69.6 \\
24 & GTM - RWPE & SLCP-DD & ChemBL & 68.7 & 58.7 & 72.8 & 58.3 & 74.3 & 73.3 & 76.2 & 61.9 & 68.0 \\
25 & GTM - RWPE & SLCP-ALL & ChemBL & 59.6 & 55.6 & 65.4 & 55.1 & 69.0 & 65.6 & 73.8 & 60.9 & 63.1 \\
26 & GTM - RWPE & SLBA & ChemBL & 62.4 & 63.0 & 63.3 & 57.2 & 59.9 & 60.8 & 73.9 & 57.0 & 62.2 \\
27 & GTL - LapPE & SLCP-DD & ChemBL & 58.3 & 55.9 & 58.8 & 54.5 & 59.0 & 61.0 & 73.8 & 57.4 & 59.9 \\
28 & GTL - LapPE & SLCP-ALL & ChemBL & 60.8 & 56.1 & 58.7 & 52.0 & 66.7 & 56.0 & 62.3 & 53.6 & 58.3 \\
29 & GTL - LapPE & SLBA & ChemBL & 61.5 & 51.8 & 56.2 & 54.5 & 60.4 & 55.1 & 73.4 & 52.8 & 58.2 \\
30 & GIN & SLCP-DD & ZINC 1M & 67.7 & 72.9 & 79.3 & 60.1 & 79.6 & 76.1 & 81.9 & 64.1 & \textbf{72.7} \\
31 & GIN & SLCP-DD & ZINC 250k & 65.6 & 75.1 & 79.0 & 59.3 & 77.2 & 76.3 & 79.2 & 64.3 & 72.0 \\
\midrule
32 & GTM - RWPE & DGI+0.5SL & ZINC 250K & 64.1 & 53.2 & 60.4 & 57.3 & 61.5 & 62.7 & 71.0 & 58.2 & 61.1 \\
33 & GTM - RWPE & DGI+1SL & ZINC 250K & 61.4 & 50.7 & 60.5 & 53.6 & 64.2 & 63.6 & 72.0 & 61.3 & 60.9 \\
34 & GTM - RWPE & DGI+2SL & ZINC 250K & 66.0 & 54.8 & 65.0 & 55.3 & 64.3 & 65.6 & 65.5 & 58.1 & 61.8 \\
35 & GTM - RWPE & DGI-SL & ZINC 250K & 46.7 & 58.2 & 64.2 & 54.7 & 63.0 & 61.2 & 69.1 & 54.0 & 58.9 \\
36 & GTL - LapPE & DGI+0.5SL & ZINC 250K & 47.6 & 50.6 & 61.4 & 53.0 & 57.6 & 59.5 & 62.1 & 56.3 & 56.0 \\
37 & GTL - LapPE & DGI+1SL & ZINC 250K & 69.4 & 55.6 & 62.1 & 56.8 & 61.5 & 59.4 & 73.0 & 57.6 & 61.9 \\
38 & GTL - LapPE & DGI+2SL & ZINC 250K & 55.3 & 50.2 & 61.8 & 52.0 & 62.9 & 58.9 & 67.1 & 55.4 & 57.9 \\
39 & GTL - LapPE & DGI-SL & ZINC 250K & 59.7 & 53.1 & 62.2 & 56.5 & 62.0 & 66.5 & 74.9 & 57.4 & 61.6 \\
40 & GIN & MASK-SL & ZINC 1M & 65.6 & 75.2 & 80.1 & 59.5 & 78.6 & 77.6 & 81.0 & 63.9 & \textbf{72.7} \\
41 & GIN & MASK-SL & ZINC 1M & 67.5 & 74.1 & 77.8 & 62.6 & 77.5 & 77.6 & 77.3 & 65.6 & \textbf{72.7} \\
\bottomrule
\end{tabular}
\caption{Test ROC-AUC (\%) performance on molecular prediction benchmarks using different pre-training strategies with GT and GIN. The rightmost column averages the mean of test performance across the 8 datasets. The best results for GT and GIN are bolded. The experiments are categorized as follows: experiments 1-8 involve direct training and evaluation of GNN models without any pre-training; experiments 9-13 utilize self-supervised pre-training; experiments 14-31 employ supervised pre-training; and experiments 32-41 utilize combined pre-training. For detailed explanations of the acronyms used, please refer to Section~\ref{sec:strategies}.}
\label{tab:results}
\end{table*}

\newpage

\section{Conclusion}
In this study, we conducted comprehensive experiments to investigate the optimal pre-training strategies for graph transformers. Our results indicate that graph transformers benefit from pre-training, particularly when using a large dataset. Specifically, supervised pre-training using computed properties offers the greatest performance gain compared to other pre-training methods. This strategy has significant potential applications in the field of biochemistry, given the availability of vast unlabeled molecule datasets \cite{ref14}, \cite{ref7} and various tools for obtaining computed properties. However, further research is necessary to identify the most effective targets for pre-training the graph transformer model.

Due to computational constraints, we did not explore other graph transformer models or investigate domains beyond biochemistry. Future work should examine the applicability of our observations to other graph transformer models and domains.

\begin{acks}
XB is supported by NUS Grant ID R-252-000-B97-133.
\end{acks}


\end{document}